\documentclass[letterpaper,10 pt,conference]{ieeeconf}  

\IEEEoverridecommandlockouts
\usepackage{cite}
\usepackage{mathptmx}
\usepackage{amsmath}
\usepackage{amssymb}
\usepackage{amsthm}
\usepackage{algorithmic}
\usepackage{graphicx}
\usepackage{hyperref}
\usepackage{cleveref}
\usepackage{textcomp}
\usepackage{xcolor}
\usepackage{wrapfig}

\def\BibTeX{{\rm B\kern-.05em{\sc i\kern-.025em b}\kern-.08em
    T\kern-.1667em\lower.7ex\hbox{E}\kern-.125emX}}

\newtheorem{theorem}{Theorem}
\newtheorem{proposition}[theorem]{Proposition}
\newtheorem{remark}[theorem]{Remark}

\newcommand{\ours}{SNDS}

\newcommand{\hc}[1]{ {\small \color{blue} HC: #1}}
\newcommand{\am}[1]{{\small \color{orange} AM: #1}}

\renewcommand{\hc}[1]{}
\renewcommand{\am}[1]{}

\begin{document}

\title{\LARGE \bf
Globally Stable Neural Imitation Policies 
}

\author{Amin Abyaneh$^{1}$, Mariana Sosa Guzmán$^{2}$, and Hsiu-Chin Lin$^{3}$
\thanks{$^{1}$Amin Abyaneh is with the Department of Electrical and Computer Engineering,
        McGill University, Montreal, Canada, 
        {\tt\small amin.abyaneh@mail.mcgill.ca}}%
\thanks{$^{2}$Mariana Sosa Guzmán is with the Department of Electrical and Computer Engineering, 
        Universidad Veracruzana, 
        Veracruz, Mexico., 
        }
\thanks{$^{3}$Hsiu-Chin Lin is with the School of Computer Science and the Department of Electrical and Computer Engineering, 
        McGill University, Montreal, Canada, 
        {\tt\small hsiu-chin.lin@cs.mcgill.ca}}
\thanks{This work is sponsored by NSERC Discovery Grant and Mitacs Globalink Research Internship.}
}

\maketitle
\thispagestyle{empty}
\pagestyle{empty}

\begin{abstract}
Imitation learning mitigates the resource-intensive nature of learning policies from scratch by mimicking expert behavior. While existing methods can accurately replicate expert demonstrations, they often exhibit unpredictability in unexplored regions of the state space, thereby raising major safety concerns when facing perturbations. 
We propose \ours{}, an imitation learning approach aimed at efficient training of scalable neural policies while formally ensuring global stability. \ours{} leverages a neural architecture that enables the joint training of the policy and its associated Lyapunov candidate to ensure global stability throughout the learning process. 
We validate our approach through extensive simulations and deploy the trained policies on a real-world manipulator arm. The results confirm \ours{}'s ability to address instability, accuracy, and computational intensity challenges highlighted in the literature, positioning it as a promising solution for scalable and stable policy learning in complex environments.
\end{abstract}

\newcommand{\stateDim}{n}
\newcommand{\stateVar}{{x}}
\newcommand{\actionVar}{\dot{\stateVar}}
\newcommand{\state}{\mathbf{\stateVar}}
\newcommand{\action}{\dot{\state}}
\newcommand{\realSpace}{\mathbb{R}^\stateDim}
\newcommand{\stateSpace}{\mathcal{X}}
\newcommand{\actionSpace}{\mathcal{A}}

\newcommand{\dataset}{\mathbf{D}}
\newcommand{\sample}{s}
\newcommand{\demonstration}{d}
\newcommand{\datasetCount}{N_t}
\newcommand{\sampleCount}{N_{\sample}}
\newcommand{\demonstrationCount}{N_{\demonstration}}
\newcommand{\sampleSet}{[\sampleCount]}
\newcommand{\demonstrationSet}{[\demonstrationCount]}
\newcommand{\stateDimensionSet}{[\stateDim]}
\newcommand{\goal}{\state^*}

\newcommand{\lpfNeuralNetSym}{v}
\newcommand{\lpfNeuralNet}{\lpfNeuralNetSym(\state)}

\newcommand{\icnnNeuralNet}{\hat{\lpfNeuralNetSym}(\state)}
\newcommand{\icnnNeuralNetSym}{\hat{\lpfNeuralNetSym}}

\newcommand{\dsFunction}{f}
\newcommand{\dsParam}{p}
\newcommand{\ds}{\hat{\dsFunction}(\state)}
\newcommand{\dsStableSym}{\pi_\theta}
\newcommand{\dsStable}{\dsStableSym(\state)}

\newcommand{\dsUnStableSym}{\hat{\pi}}
\newcommand{\dsUnStable}{\dsUnStableSym(\state)}

\newcommand{\lpfFunction}{v}
\newcommand{\lpfParam}{q}
\newcommand{\lpf}{\lpfFunction(\state)}

\newcommand{\lossSym}{\mathcal{L}}

\section{Introduction}
\label{sec:introduction}

\noindent
Imitation learning (IL) is a pivotal notion attempting to overcome the diverse safety and complexity challenges of policy learning by emulating expert behavior~\cite{schaal1999imitation,hussein2017imitation}. IL facilitates the training of intricate motion policies without resorting to an exhaustive search in the robot's state space~\cite{figueroa2022locally,ds_billard}. Nonetheless, only a handful of IL methods offer formal guarantees pertaining to the stability of the resulting policies. Global stability assumes a critical role when deploying the policy in stochastic environments prone to external perturbations. Notably, it assures that the policy can recover effectively and \emph{predictably} to a predetermined target, even in uncharted regions of the state space that lie beyond the scope of expert demonstrations.

\begin{figure}[ht]
    \centering
    \includegraphics[width=\linewidth]{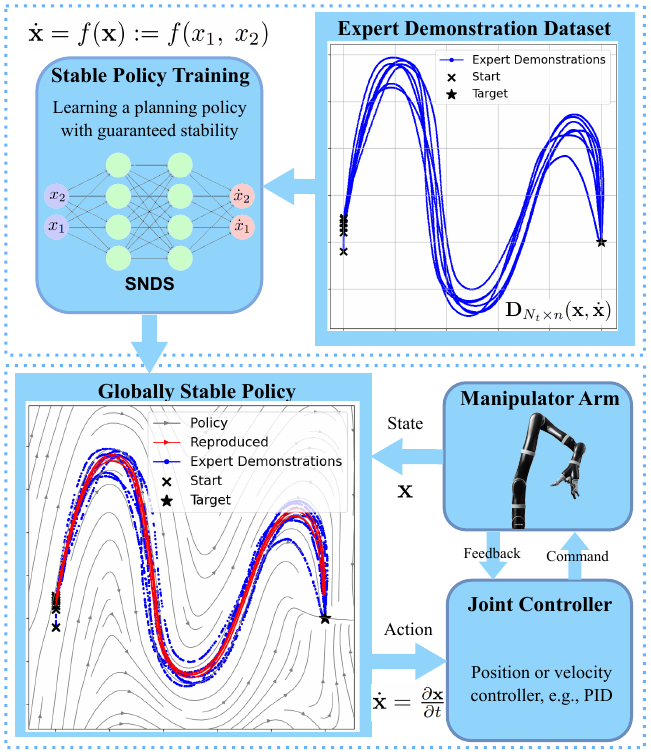}
    \caption{Overview of the proposed stable neural policy learning method. Policy learning (top) optimizes a Lyapunov-stable neural policy over the expert demonstration data. The optimized policy is then deployed (bottom) to plan globally stable trajectories resistant to unpredictable perturbations.}
    \label{fig:snds_overview}
\end{figure}

Prior research in safe IL is focused on autonomous and time-invariant dynamical systems to model and train motion policies~\cite{ds_billard}. These policies are optimized across a set of expert trajectories, yielding a proper action (velocity) contingent on the current state (position). The optimization process ensures global stability through established theoretical frameworks, such as Lyapunov or contraction theories~\cite{khansari2011learning, khansari2014learning, figueroa18a, neumann2015learning, ravichandar2017learning, khadivar2021learning, abyaneh2023plyds}. Nevertheless, reproducing complex trajectories in high-dimensional state spaces with the previous methods can be impractical due to computational complexity, non-convex optimization, and sample inefficiency. To address computational cost, previous research~\cite{figueroa18a, ravichandar2017learning, abyaneh2023plyds} tends to restrict the class of functions of the contraction metric or the Lyapunov candidate, which in turn limits the imitation performance of the optimized policy. Most notably, a common limitation of these methods appears in the absence of neural policy representation, which are recognized for their scalability, efficient gradient-based optimization, and domain transfer capabilities~\cite{chang2019neurallyapunov}.

Recently developed stable neural policies \cite{rana2020euclideanizing, zhang2022riemaniandiffeomorphism} mainly employ diffeomorphism and invertible mappings to transform a simple stable policy into a highly nonlinear one. Yet, the training stage requires multiple demonstrations, leads to policies with quasi-stability, and hinders incremental learning for slightly different expert trajectories. Neural IL methods, on the other hand, offer flexible and precise policies~\cite{ho2016generative, airl_imitation, dai2021lyapunov, coulombe2022}, but lack the required safety and reliability outside the region covered by expert demonstrations and are often data hungry. Similar limitations persist in inverse reinforcement learning~\cite{abbeel2004apprenticeship, ziebart2008maximumentropy}, where computational complexity is fueled by having reinforcement learning in an inner loop~\cite{ho2016generative}.

To tackle these challenges, we represent policies with \textbf{S}table \textbf{N}eural \textbf{D}ynamical \textbf{S}ystems (\ours{}), where we jointly train a neural dynamical system policy, alongside a secondary, convex-by-design network that guarantees global stability. 
Subsequently, a joint optimization process trains the neural policy by minimizing a novel differentiable loss aimed at strengthening the alignment between policy rollouts and expert demonstrations. \ours{}, therefore, benefits from an expressive and stable neural representation, allowing for safe and scalable approximation of the underlying dynamical system. We present an overview of our framework in \Cref{fig:snds_overview}, and outline our key \textbf{contributions} below. 
\begin{itemize}
    \item Designing \ours{}—a {\it stable-by-design} neural representation for nonlinear dynamical systems as motion policies based on expert demonstrations. 
    \item Providing formal stability analysis founded on Lyapunov theory and convex neural networks.
    \item Formulating a differentiable trajectory alignment loss function, inspired by forward Euler's method.     
    \item Empirical evaluation of \ours{}'s effectiveness in higher state space dimensions for complex trajectories, both in simulation and real-world scenarios. 
\end{itemize}


\section{Background}
\label{sec:background}

\subsection{Preliminaries}
\label{sec:preliminaries}

\noindent \textbf{Dynamical system. } A dynamical system (DS) describes the evolution of a state, $\state$, in the ambient space $\stateSpace \subset \mathbb{R}^n$, over time~\cite{devaney2021introduction}. We consider an autonomous, time-invariant DS modeled with a first-order ordinary differential equation, $\action = \dsFunction(\state)$, where the system yields the time derivative for each $\state$, without any control input and independent of time. 
\vspace{0.1cm}

\noindent \textbf{Lyapunov stability theory. } $\dsFunction(\state)$ is globally asymptotically stable (GAS) at an equilibrium, $\goal$, if for any initial state, the system approaches $\goal$ as time progresses towards infinity. The Lyapunov stability theory is a widely used tool to analyze the GAS property of dynamical systems. 
According to the theory, a system exhibits GAS if there exists a positive-definite function $\lpfNeuralNet: \stateSpace \xrightarrow{} \mathbb{R}$, referred to as the Lyapunov potential function (LPF), satisfying $\dot{\lpfNeuralNetSym}(\state) < 0$ 
for all $\state \neq \goal$, and $\dot{\lpfNeuralNetSym}(\goal) = 0$ at the equilibrium.
$$
\forall \state_0 \in \stateSpace, \;\; \state_{t} = \dsStableSym(\state_{t-1}, \mathbf{a}), \;\; \lim_{t \rightarrow \infty} \; \state_t = \state^* 
$$
\vspace{0.1cm}

\noindent \textbf{Input convex neural networks (ICNN). } ICNN's special architecture enables universal approximation of convex functions while maintaining convexity during training \cite{icnn_paper,manek2019learning}. 
An $L$-layer ICNN, $\icnnNeuralNetSym$, is formulated as follows for $l \in [L]$:
\begin{gather*}
    \icnnNeuralNet = z_L, \quad z_{l+1} = \sigma_l( U_l x + W_l z_l + b_l),
\end{gather*}
where $z_l$ denotes the layer output, and $W_l$ and $b_l$ are real-value weights. In contrast to feedforward networks, ICNN design exploits $U_l$, that are positive weights connecting the input directly to each layer, and $\sigma_l$ can be any convex, non-decreasing activation function. 

\subsection{Problem statement}
\label{sec:problem}
\noindent 
 Consider a policy that functions within a state space denoted by $\stateSpace \subset \realSpace$. The state space may correspond to a robot's task ($\mathit{T}$) or configuration ($\mathit{C}$) space. The policy outputs an action in $\actionSpace \subset \realSpace$, such as velocity or torque, which determines the change in the state over time. 
We denote the state variable by $\state \triangleq [\stateVar_1 ; \stateVar_2 ; \ldots ; \stateVar_\stateDim]^T \in \stateSpace$, and assume that the yielded action indicates the time derivative of the current state, denoted by $\action = \frac{\partial x}{\partial t} \in \actionSpace$. In this context, our \textbf{primary objective} is to learn a globally stable imitation policy to map $\state$ to $\action$, provided a dataset of expert's state-action pairs.

Let $\demonstrationCount \in \mathbb{N}$ be the number of trajectories demonstrated by the expert. Each trajectory contains $\sampleCount \in \mathbb{N}$ state-actions. The dataset of expert trajectories stacks all state-action pairs,
\begin{gather} \label{eq:dataset_def} \dataset \triangleq \Bigl\{\big(\state^{\demonstration}[\sample], \; {\mathbf{a}}^{\demonstration}[\sample]\big) \;\big|\; \demonstration \in \demonstrationSet,\; \sample \in \sampleSet\Bigr\},
\end{gather}
where $(\state^{\demonstration}[\sample], \; \action^{\demonstration}[\sample])$ is the dataset entry corresponding to the $\sample$-th sample of the $\demonstration$-th trajectory. The dataset $\dataset$ holds $\datasetCount = \demonstrationCount\sampleCount$ samples. We assume that the trajectories share a common target endpoint, $\state^* \in \stateSpace$, and have zero velocity at the target, i.e., $\state^{\demonstration}[\sampleCount] = \state^*$ and $\action^{\demonstration}[\sampleCount] = \mathbf{0}, \; \forall \demonstration \in \demonstrationSet$.

The mapping between states and actions can be modeled with a time-invariant autonomous DS~\cite{khansari2011learning}, denoted by:
\begin{gather} \label{eq:main_ds_def}
    \dsStableSym(\textbf{a} \mid \textbf{x}), \;\;  \dsStableSym: \stateSpace \xrightarrow{} \actionSpace, \; \text{s.t.}, \; \textbf{a} = \frac{\partial \textbf{x}}{\partial t}
\end{gather}

\noindent 
In \Cref{eq:main_ds_def}, the function $\dsFunction$ corresponds to an ordinary differential equation describing the true underlying DS. The term $\epsilon \in \realSpace$ accounts for the effect of measurement noise present in the expert's demonstrations, which is incorporated into the estimated DS. Our subsequent objective is to acquire a noise-free approximation of $\dsFunction(\state)$, denoted as $\dsStable$. This estimated function,  $\dsStable$, can be perceived as a \textit{policy} that maps states to actions in real-time, thereby guiding the robot with the right set of actions to imitate the demonstrated trajectories within the state space. 

Alternatively, a trajectory can be generated through forward Euler's method. Imagine a robot situated at $\state[0]$, the policy generates an action $\action[0] = \dsStableSym(\state[0])$, and the next state is subsequently calculated through $\state[1] = \state[0] + \Delta t \dsStableSym(\state[0])$, where $\Delta t$ determines the granularity of the discretization. 

\section{Methodology}
\label{sec:methodology}
\noindent 
We model the policy, $\dsStable$, and the corresponding LPF, $\lpfNeuralNet$, with two neural networks in \Cref{sec:policy_formulation}. This allows us to accurately imitate an expert's behavior, while providing formal GAS analysis in \Cref{sec:global_stability}. Lastly, we introduce a differentiable loss function in \Cref{sec:srvf_training_loss} to improve both sample efficiency and imitation accuracy of \ours{}.

\subsection{Dynamical system policy formulation}
\label{sec:policy_formulation}

\begin{figure}[t]
    \centering
    \includegraphics[width=0.99\linewidth]{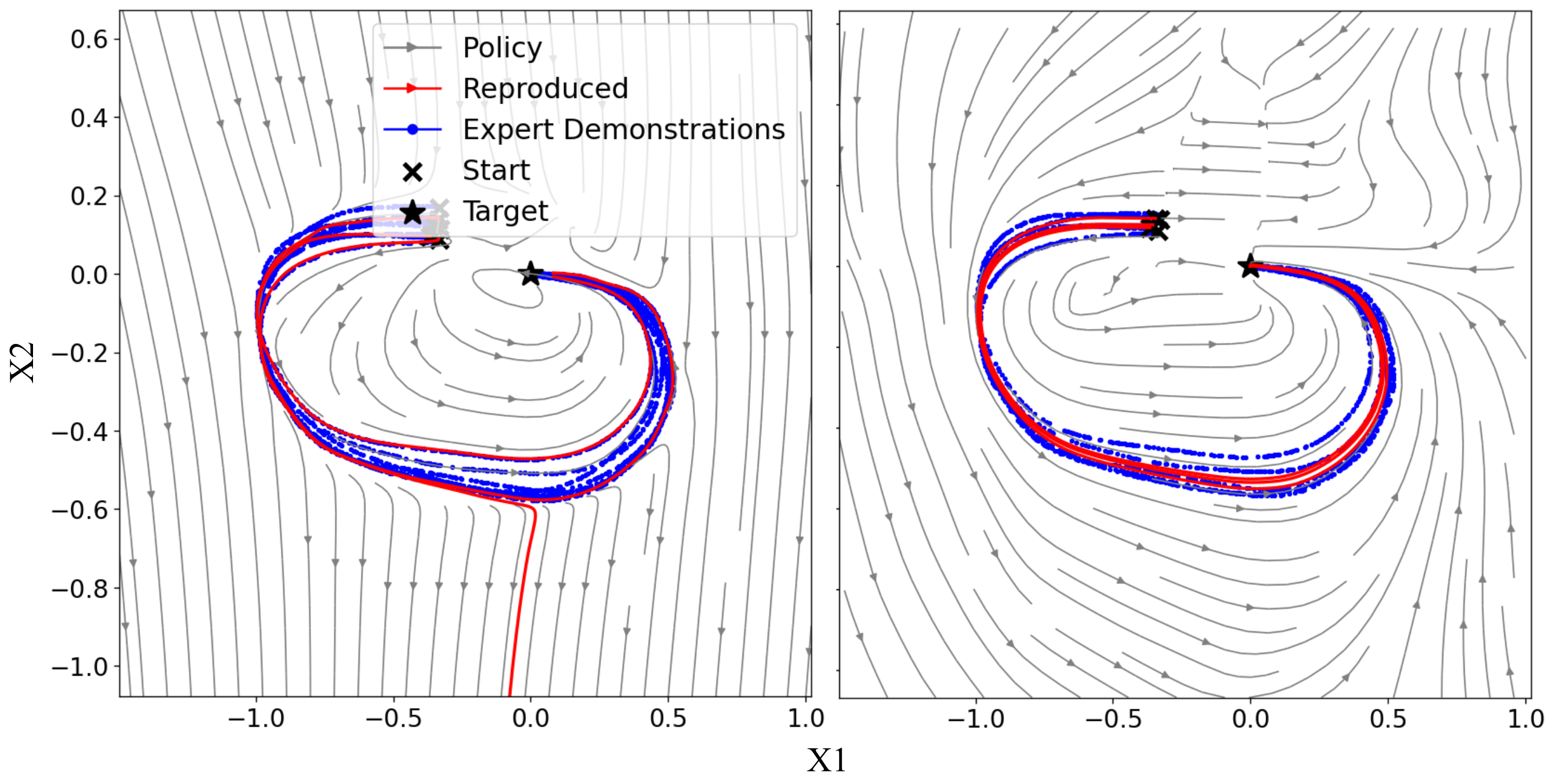}
    \caption{An example of unstable (left) vs. a stable (right) policies optimized on expert's data from the handwriting dataset \cite{handwriting_lasa}. While both policy rollouts can reproduce the expert motion, an unstable policy cannot recover from perturbation that push the robot to unknown state space regions.}
    \label{fig:ablation_rollout}
\end{figure}

\noindent 
Let $\dsUnStable: \stateSpace \rightarrow \actionSpace$ be a standard and unrestricted feedforward neural network which models the unstable policy. Although the model can capture intricate expert demonstrations, training $\dsUnStable$ solely on $\dataset$ results in an unstable policy which cannot predictably recover even from mild perturbations, as depicted in \Cref{fig:ablation_rollout}. Hence, in this section, we enforce the trained policy to have the GAS property by satisfying the conditions of Lyapunov stability theory in \Cref{sec:preliminaries}. 

The first step is to define a positive-definite and differentiable LPF. As mentioned in Section~\ref{sec:preliminaries}, ICNNs can approximate any convex function.
To ensure that the LPF is positive definite, we employ a well-studied technique \cite{dai2021lyapunov, coulombe2022} to define it as, 
\begin{gather}
\label{eq:lpf_definition}
\lpfNeuralNet = \big[\icnnNeuralNetSym(\state) - \icnnNeuralNetSym(\goal)\big] + \delta|| \state - \goal||_2^2,
\end{gather}
where  $\lpfNeuralNet: \stateSpace \rightarrow \mathbb{R}$ is still a convex function. The first term ensures $\lpfNeuralNetSym(\goal) = 0$, while the addition of the state's ${l}_2$-norm with a negligible $\delta > 0$, along with convexity of ICNN guarantee that $\lpfNeuralNet > 0, \; \forall \; \state \neq \goal, \; \state \in \stateSpace$. Hence, the LPF with this architecture, $\lpfNeuralNet$, satisfies the positive-definite condition required by Lyapunov stability theory.

Next, to satisfy the negative derivative condition of Lyapunov theory, i.e., $\frac{d\lpfNeuralNetSym(\state)}{dt} =  \nabla \lpfNeuralNet^T \dsUnStable < 0$, we modify the projection expression in \cite{manek2019learning} to enforce GAS while training $\dsUnStable$. The policy $\dsStable$ resulting from projecting $\dsUnStable$ onto the half space, $\nabla \lpfNeuralNet^T \dsUnStable < 0$, 
is formulated as follows, 

\begin{equation}
\label{eq:projection_formula}
\begin{aligned}    
    \dsStable = \dsUnStable - \nabla \lpfNeuralNet \frac{\sigma(\nabla \lpfNeuralNet^T \dsUnStable)}{\lVert \nabla \lpfNeuralNet \rVert_2^2}.
\end{aligned}
\end{equation}  

\noindent
In \Cref{eq:projection_formula}, if the network output $\dsUnStable$ fails to meet the Lyapunov condition, i.e., $ \nabla \lpfNeuralNet^T \dsUnStable >= 0$, the output is projected so that $\dsStable$ always fulfills the condition. The process is simplified with a ReLU activation function. Note that the original formulation in \cite{manek2019learning} enforces exponential stability, which is too restrictive for our problem. Hence, we further relax the projection to guarantee GAS. 

\begin{figure}[t]
    \centering
    \includegraphics[width=1\linewidth]{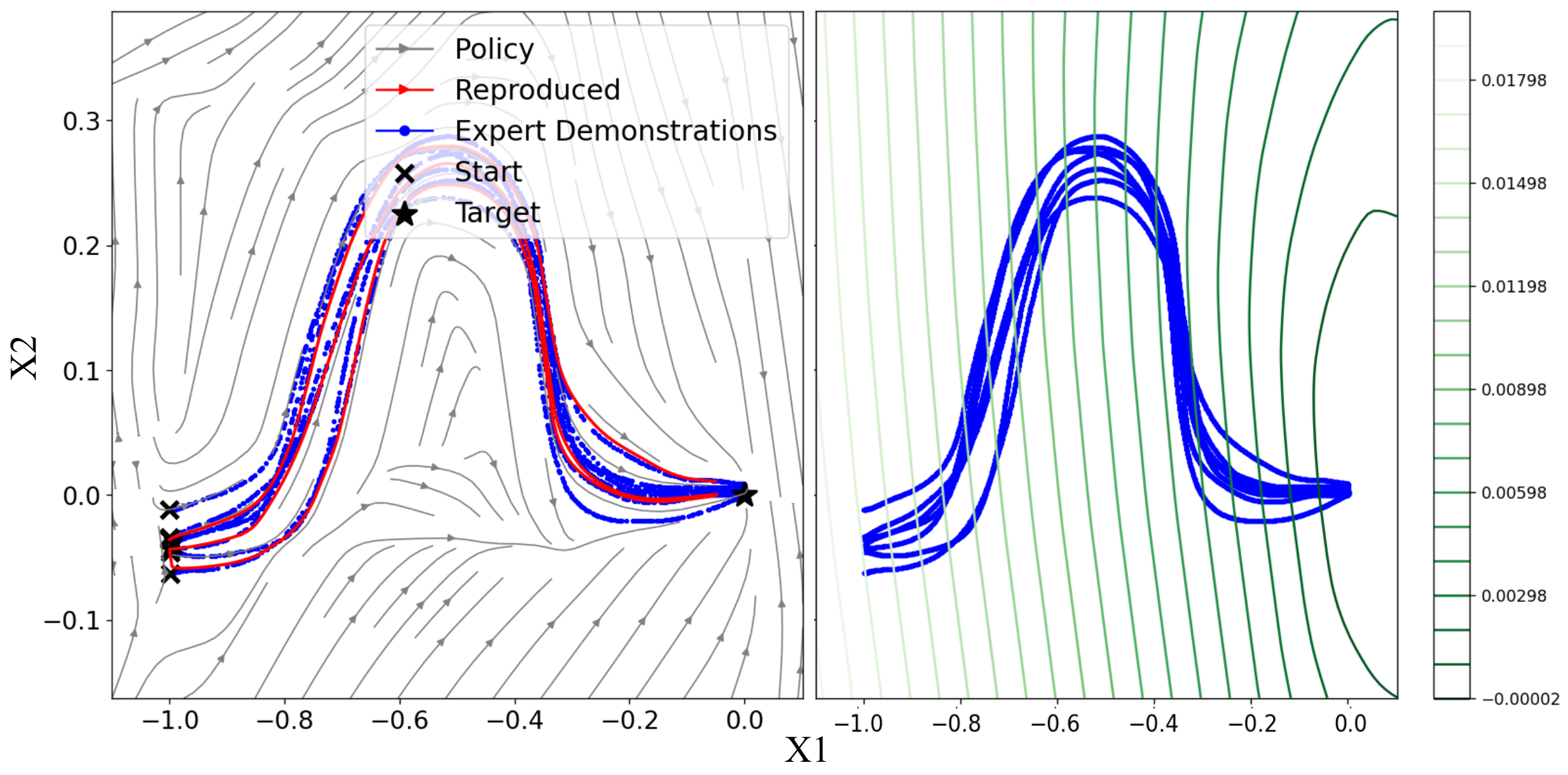}
    \caption{Joint training of globally stable neural policies (left) and the corresponding LPF (right) optimized on expert's data from the handwriting dataset \cite{handwriting_lasa}.     
    The contours for the LPF (green) illustrate both positive-definiteness and convexity of the trained function. 
    }
    \label{fig:rollout_lpf_contours}
\end{figure}

 \Cref{fig:rollout_lpf_contours} portrays an example of trained $\dsStable$ and $\lpfNeuralNet$, and the resulting global stability. The differentiable and globally stable $\dsStable$, can be trained using efficient gradient-based methods within its parameters, as long as $\dsUnStable$ and $\lpfNeuralNet$ are defined using automatic differentiation tools. 

\subsection{Global asymptotic stability guarantees}
\label{sec:global_stability}
\noindent
To establish that $\dsStable$ defines a function with GAS property, we use proof by contradiction to show that regardless of initial state, $\state[0]$, every trajectory converges to the target, that is, $\lim_{k \rightarrow \infty} \state[k] = \goal$.

\begin{proposition}
\label{stability_proposition}
The dynamical system, $\dsStable$, in \Cref{eq:projection_formula} is globally asymptotically stable with the Lyapunov function, $\lpfNeuralNet$, defined in \Cref{eq:lpf_definition}, and any two arbitrary networks, $\dsUnStable$ and $\icnnNeuralNet$, with bounded real-valued weights.
\end{proposition}
\prooff We leverage proof by contradiction for this proposition. 
Let $\{\state[0], \ldots, \state[k]\}$ be a trajectory not converging to the target, meaning that $\lim_{k\rightarrow \infty} \state[k] \neq \goal$. Since $\lpfNeuralNet$ is decreasing, as formulated in \Cref{eq:projection_formula}, and $\lpfNeuralNet \geq 0$, as defined in \Cref{eq:lpf_definition}, there must exist a value $\mu \in\mathbb{R}^+$ such that $\lim_{k\rightarrow \infty} \lpfNeuralNetSym(\state[k])=\mu$. We reason that $\mu \leq \lpfNeuralNetSym(\state[i]) \leq \lpfNeuralNetSym(\state[0])$ for every $0 \leq i \leq k$. 

Now consider the set $\mathbf{S}$ of all $\state[i], \forall 0 \leq i \leq k$. This set is \emph{compact}. More precisely, for any open cover~\footnote{An open cover for a set is a collection of open sets whose union contains the original set.} of that set, there exists a finite subcover, which means that a finite number of the open sets from the cover are sufficient to cover the original set. Hence, $\dot{\lpfNeuralNetSym}(\state)$ takes its supremum, $\sup_{\mathbf{S}} \dot{\lpfNeuralNetSym}(\state) = -\alpha$ and $\alpha \in \mathbb{R}^+$ over this set. This conclusion is justified based on the projection introduced in \Cref{eq:projection_formula}. The core integration in the proof of Lyapunov theorem,
\begin{gather*}
 \lpfNeuralNetSym(\state[T]) = \lpfNeuralNetSym(\state[0]) + \int_0^T \dot{\lpfNeuralNetSym}(\state[t])dt \leq \lpfNeuralNetSym(\state[0]) - \alpha T, 
\end{gather*}
indicates a \emph{contradiction} when, $T = \frac{\lpfNeuralNetSym(\state[0])}{\alpha} + \beta$ and $\beta \in \mathbb{R}^+$: 
\begin{gather*}
    \lpfNeuralNetSym(\state[T]) \leq \lpfNeuralNetSym(\state[0]) - \alpha (\frac{\lpfNeuralNetSym(\state[0])}{\alpha} + \beta) = - \alpha \beta < 0,
\end{gather*}
The above equation contradicts the fact that $\lpfNeuralNet > 0$. Hence, all trajectories must converge to the desired target, $\goal$. \hfill $\square$

\begin{remark}
\Cref{stability_proposition} only requires $\lpfNeuralNet$ to be convex. The policy network, $\dsStable$, remains fairly unrestricted to capture intricate expert demonstrations. 
\end{remark} 

\noindent
Thus, the proposed \ours{} formally establishes global stability using an efficient gradient-based search for the LPF within the class of convex functions.

\subsection{SRVF training loss}
\label{sec:srvf_training_loss}
\noindent When constructing an optimization loss, the literature often opts for the conventional Mean Squared Error (MSE) \cite{khansari2011learning, rana2020euclideanizing, figueroa2022locally} to gauge the dissimilarity between the policy's output and the actual velocity labels over each batch of data. Consequently, the MSE loss is solely focused on the precision of estimated velocity. While reproducing exact velocities is essential, even slight inaccuracies at one instance can lead to accumulated error in the reproduced trajectories. The accumulation happens as a result of using forward Euler's method that relies on previous estimates of $\state$ to yield the next ones. Therefore, a more effective loss function should promote the accuracy of trajectories generated through forward Euler's  method during policy rollouts.

Employing forward Euler's method to generate a full trajectory makes the gradient-based optimization intractable as a result of repetitive integrations. To mitigate this effect, we propose a limited horizon combination of the Square-Root Velocity Functions (SRVF)~\cite{bruveris2016srvf} and traditional MSE to design the loss function. SRVF methods involve normalizing the curve's velocity, with a focus on changes in curve shape. Assuming a trajectory $\state^d_k = \{\state^d[0], \ldots, \state^d[k]\}$, we shape the features by the definition:

\begin{gather}
\texttt{SRVF} (\state^d_k) = \sqrt{\frac{\state^d_k[s + \Delta t] - \state^d_k[s]}{\|\state^d_k[s + \Delta t] - \state^d_k[s]\|}}, \quad \Delta t = 1.
\end{gather}

\noindent
Given the state-action pairs, $\state^\demonstration[\sample], \; \action^\demonstration[\sample] \sim \texttt{SRVF} (\dataset)$, and the differentiable dynamics $\dsStable$, we define the loss $\mathcal{L}$ as,
\begin{gather}
\label{eq:srv_loss}
\begin{aligned}    
     & \lossSym(\state; \; \theta) = \; \gamma_0 \mathop{\mathbb{E}} \Big[\big(\dsStableSym(\state^{\demonstration}[\sample]) - \action^\demonstration[\sample]\big)^2\Big] \; + \\ 
     \sum_{i \in [N_w]} \gamma_i \mathop{\mathbb{E}} & \Big[\big(\state^{\demonstration}[\sample + i - 1] +  \dsStableSym (\state^{\demonstration}[\sample + i - 1]) \Delta t - \state^\demonstration[\sample + i]\big)^2\Big], 
\end{aligned}    
\end{gather}
\am{How to move the equation number to the first line to free up space?}where $\gamma_i$ are discount factors decreasing as the loss horizon expands. The intuition behind \Cref{eq:srv_loss} is rather straightforward: $\lossSym$ is a differentiable function encapsulating both the current label mismatch, and deviation from the generated trajectory for a fixed horizon of $N_w$ consecutive samples. As long as the horizon is limited, stochastic optimizers, such as ADAM~\cite{adam_optimizer}, can optimize the loss on expert's training data.

\section{Experiments}
\label{sec:experiments}
\noindent
We utilize two sets of motion planning data in our experiments. Our primary dataset is sourced from the well-known LASA Handwriting dataset \cite{handwriting_lasa}, which contains records of handwritten trajectories on a tablet. The second dataset consists of more complex motions gathered by \cite{figueroa18a} in the same way. Our real-world experiments are only based on the latter due to its complex nature.

\subsection{Evaluation}
\label{sec:evaluation}

\begin{figure}[t]
    \centering
    \includegraphics[width=\linewidth]{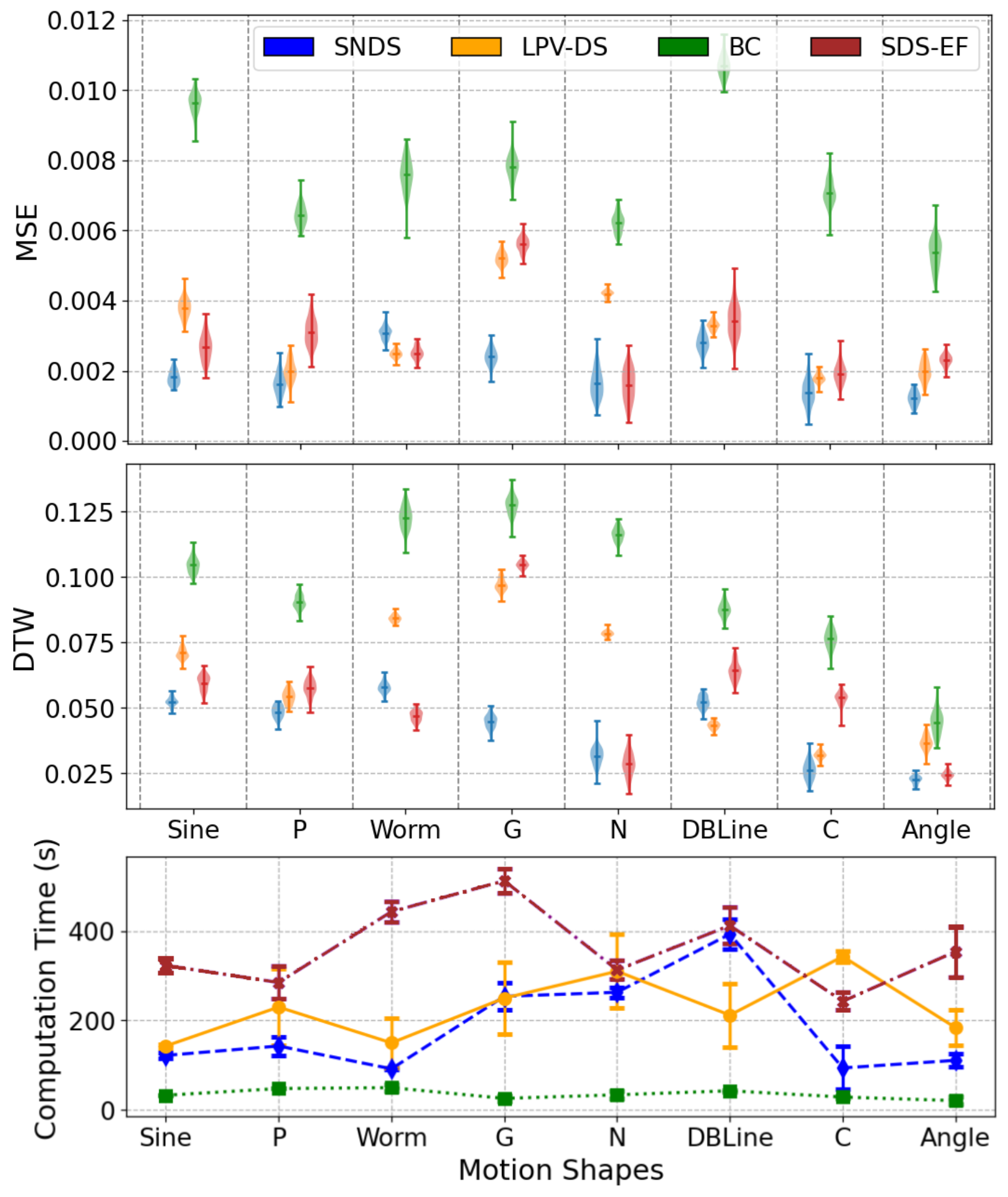}
    \caption{Comparing the reproduction accuracy (top) and computational cost (bottom) of \ours{} against baseline methods using the MSE and DTW metrics introduced in \Cref{eq:mse_metric} and \Cref{eq:dtw_metric}, respectively. The accuracy of policy actions and the learning time against other \emph{stable} methods remain comparably lower for \ours{} across most designated trajectories.} 
    \label{fig:mse_baseline}
\end{figure}
\noindent
We compare our approach against three existing baselines: Behavioral Cloning (BC) \cite{pomerleau1988behavioralcloning, torabi2018behavioral}, Linear Parameter-Varying Dynamical System (LPV-DS) \cite{figueroa18a}, and Stable Dynamical System Learning Using Euclideanizing Flows (SDS-EF) \cite{rana2020euclideanizing}. Among these methods, only BC lacks formal global asymptotic stability  guarantees. Nevertheless, comparing \ours{} against BC enables us to gauge \ours{}'s accuracy and global stability against an unrestricted neural policy.

To assess the effectiveness of \ours{} in comparison with the named baselines, we carry out the learning process for each method on designated handwriting dataset motions. For each motion, we randomly split the demonstrated dataset in the dataset into train ($0.8$) and test ($0.2$) sets. The policy learning stage is carried out on the training data, and the optimal policy is used to generate policy rollouts, $\state^g$, using the Euler's method. Policies are subsequently evaluated by calculating Mean-Squared Error (MSE),
    \begin{equation}
    \label{eq:mse_metric}
    MSE(\action^g, \action^\demonstration) =\frac{\sum_{\demonstration=1}^{\demonstrationCount^{test}} \sum_{\sample=1}^{\sampleCount} (\dsStableSym(\state^{\demonstration}[\sample]) - \dot{\state}^\demonstration[\sample])^2}{2 \demonstrationCount^{test} \sampleCount}, 
    \end{equation}
and Dynamic Time Warping (DTW),
    \begin{equation}
    \label{eq:dtw_metric}
    DTW(\state^g, \state^\demonstration) = \min_{p \in \mathcal{P}({\state}^g, {\state}^d)}
        \left( \sum_{(i, j) \in \pi} \mathit{Euc}(\state^g[i], \state^d[j])^q \right)^{\frac{1}{q}},
    \end{equation}
for handwriting dataset and experiments in higher state space dimensions to better capture discrepancies between policy rollouts and expert demonstrations. Within \Cref{eq:dtw_metric}, $p$ represents an alignment path, $\mathcal{P}$ is the set of all admissible paths, $\mathit{Euc}$ is Euclidean distance~\cite{keogh2005dtw}, and $q = 2$. Notice that to calculate DTW, we need to generate an entire trajectory, while with MSE, only the generated actions are compared against true labels. We conduct the training and assessment procedures repeatedly with 10 random seeds, and present the mean and standard deviation of the results.

\begin{figure}[t]
    \centering
    \includegraphics[width=1\linewidth]{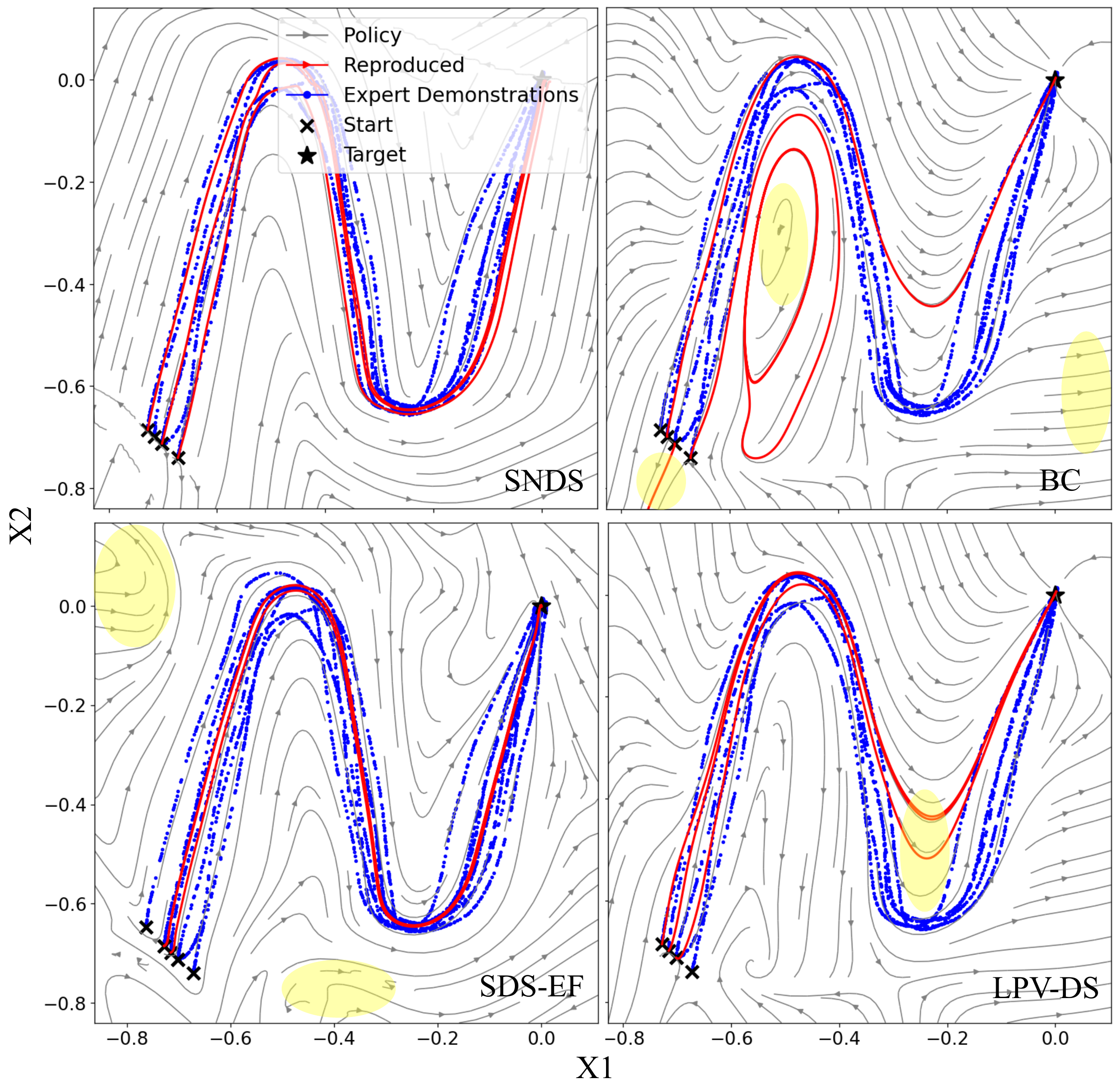}
    \caption{Policy rollouts for \ours{} and other baselines. Notice the highlighted inaccuracies or instabilities (yellow)  for other methods. The acquired policies are optimized using the N-shaped data of the handwriting dataset.}
    \label{fig:baseline_rollout}
\end{figure}

\subsection{Handwriting dataset policies}
\label{handwriting_dataset}
\noindent 
We apply \ours{} and selected baselines to each set of demonstrations as explained in \Cref{sec:evaluation}, and compare the reproduction accuracy and computation time~\footnote{Note that we restrict our computational resources to a single Core-i7 Gen8 CPU, despite that \ours{} and SDS-EF can leverage parallel computing.} across various motions in the handwriting dataset. \Cref{fig:mse_baseline} presents a numerical comparison of both reproduction accuracy and computation time between \ours{} and the selected baselines. The results indicate an acceptable level of reproduction accuracy while providing GAS certificates. In rare cases, \ours{} shows a higher error compared to baselines, but outperforms others for the majority of elected motions, and maintains a reasonable computation time.

\begin{figure}[t]
    \centering
    \includegraphics[width=1\linewidth]{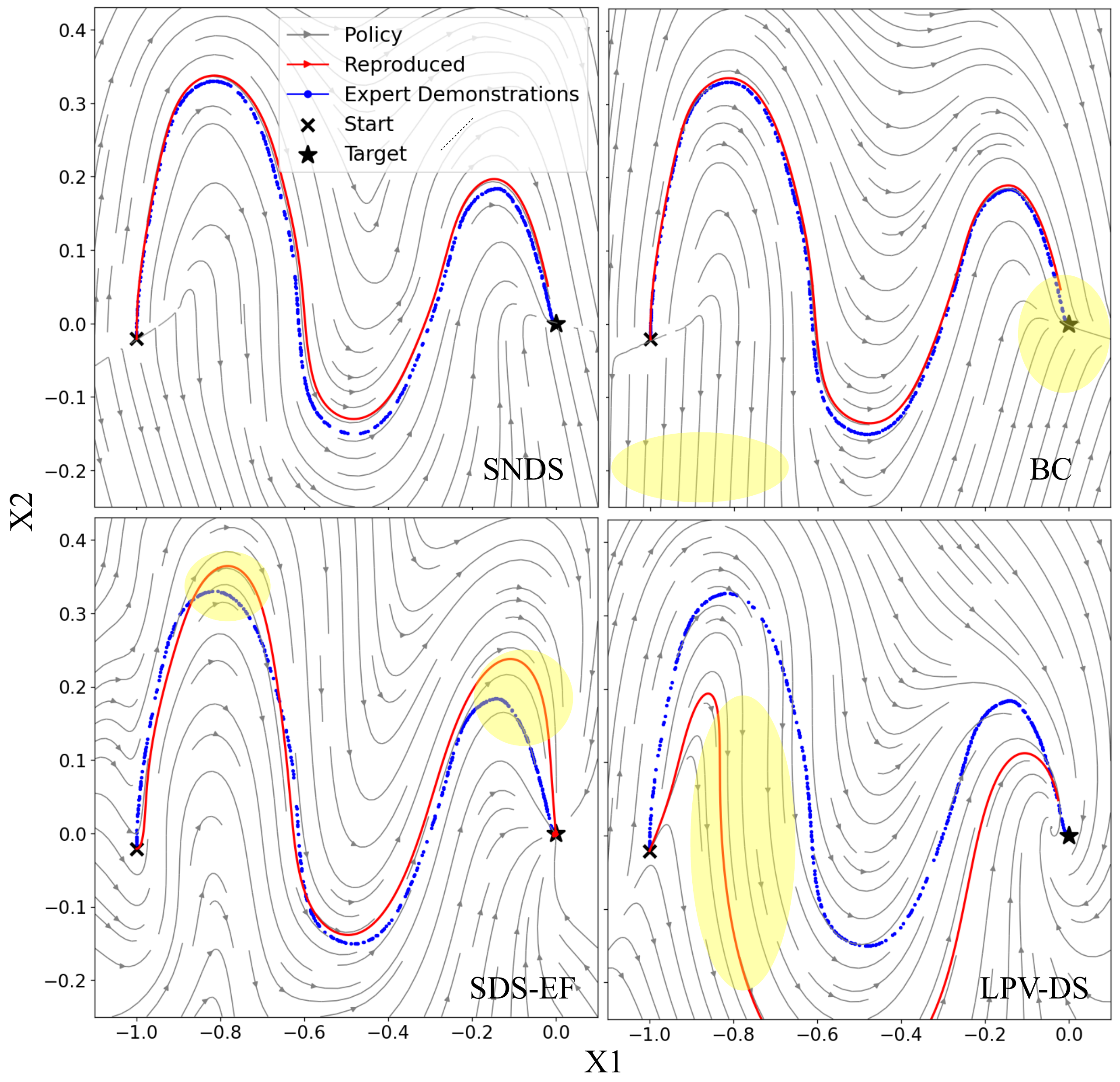}
    \caption{Policy rollouts when training on a single Sine-shaped demonstration of the handwriting data. Inaccuracies or instabilities are highlighted. \ours{} performs better owing to the expressive architecture and customized loss.}
    \label{fig:baseline_rollout_lessdem}
\end{figure}

To closely examine the GAS property of each method, we plot the acquired policies using streamlines and generated two simulated rollouts with Euler's method in \Cref{fig:baseline_rollout}. It becomes apparent that \ours{}, and LPV-DS provide predictability through guaranteeing GAS, while other approaches, such as BC and SDS-EF fail to render similar certificates. On a general note, even though we observe unstable policies trained with SDS-EF, instabilities occur only for a subset of motions in the handwriting dataset, and tend to intensify in regions further away from expert's demonstrations.

To further showcase the sample inefficiency associated with Gaussian mixture model and diffeomorphism-based methods, such as LPV-DS and SDS-EF, respectively, we repeat the experiments with the training set reduced to only {\it a single demonstration}. The learned policies in this scenario are illustrated in Figure~\ref{fig:baseline_rollout_lessdem}, indicating the sample efficiency of \ours{} when compared to LPV-DS and SDS-EF in the presence of limited demonstration samples.

\subsection{SE(3) policy training}
\label{manipulation_tasks}
\noindent 
To illustrate the applicability of our method in high-dimensional state spaces, such as SE(3), we consider both the position and orientation of the robot and redefine the state variable, $\state$, accordingly.  
We pick the \emph{snake motion}—an overly complex and nonlinear motion presented in \cite{figueroa18a} and 2 different motions in the handwriting dataset, namely \emph{Sine-} and \emph{G-shaped} motions, to consolidate our higher dimensional performances. 
We use synthetic demonstrations for orientation, as the demonstrations only comprise translation and linear velocity. After training \ours{} on this data, the policy generates trajectories in SE(3) that guide the end-effector toward the desired pose.

\begin{figure}[t]
    \centering
    \includegraphics[width=1\linewidth]{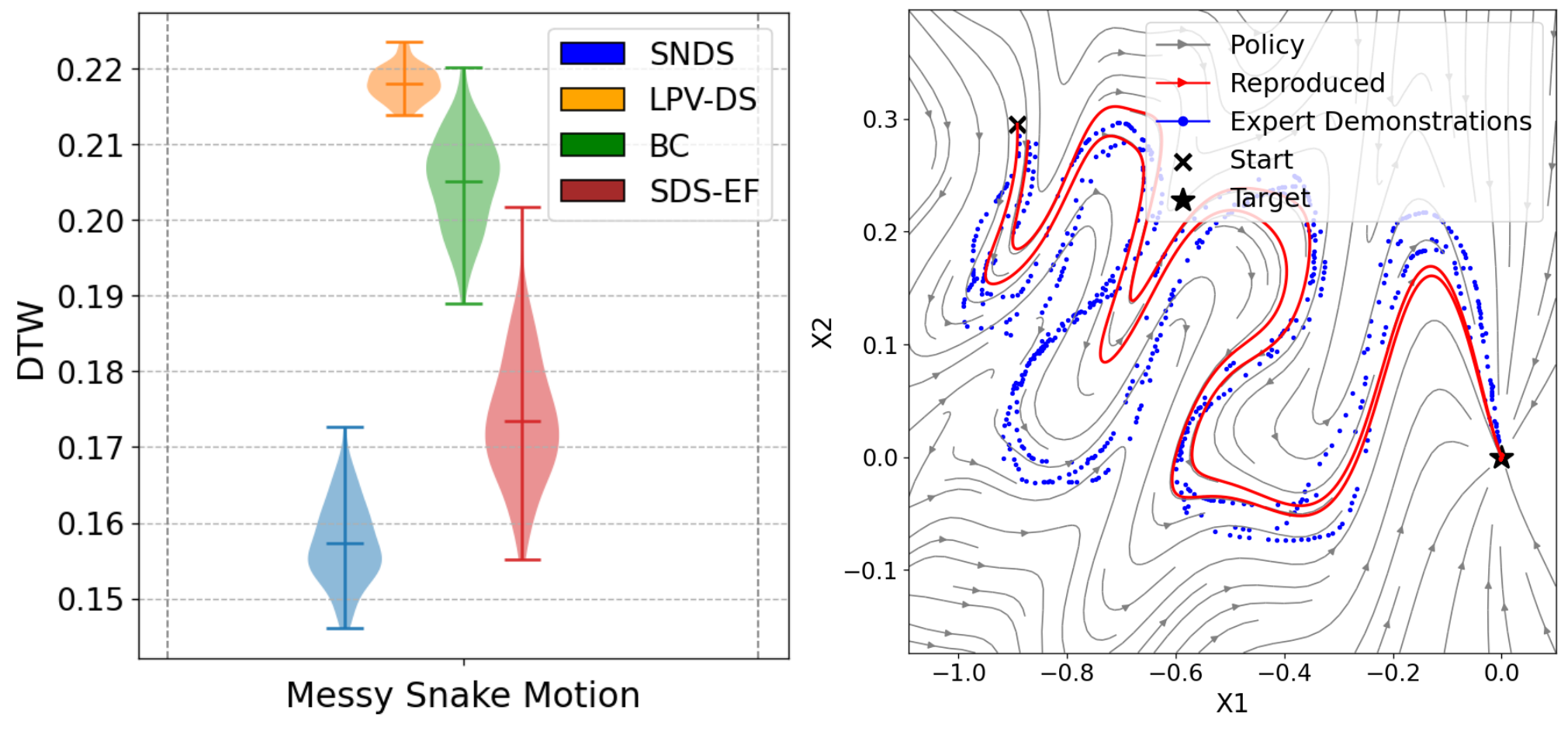}
    \caption{DTW between policy rollouts and expert's data for the complex Messy Snake demonstrations in SE(3) (left). \ours{} can better extend to higher state space dimensions compared to former baselines. The trained \ours{} policy over the expert demonstrations is depicted on the right. }
    \label{fig:rollout_messy_snake}
\end{figure}

\begin{figure}[t]
    \centering
    \includegraphics[width=1\linewidth]{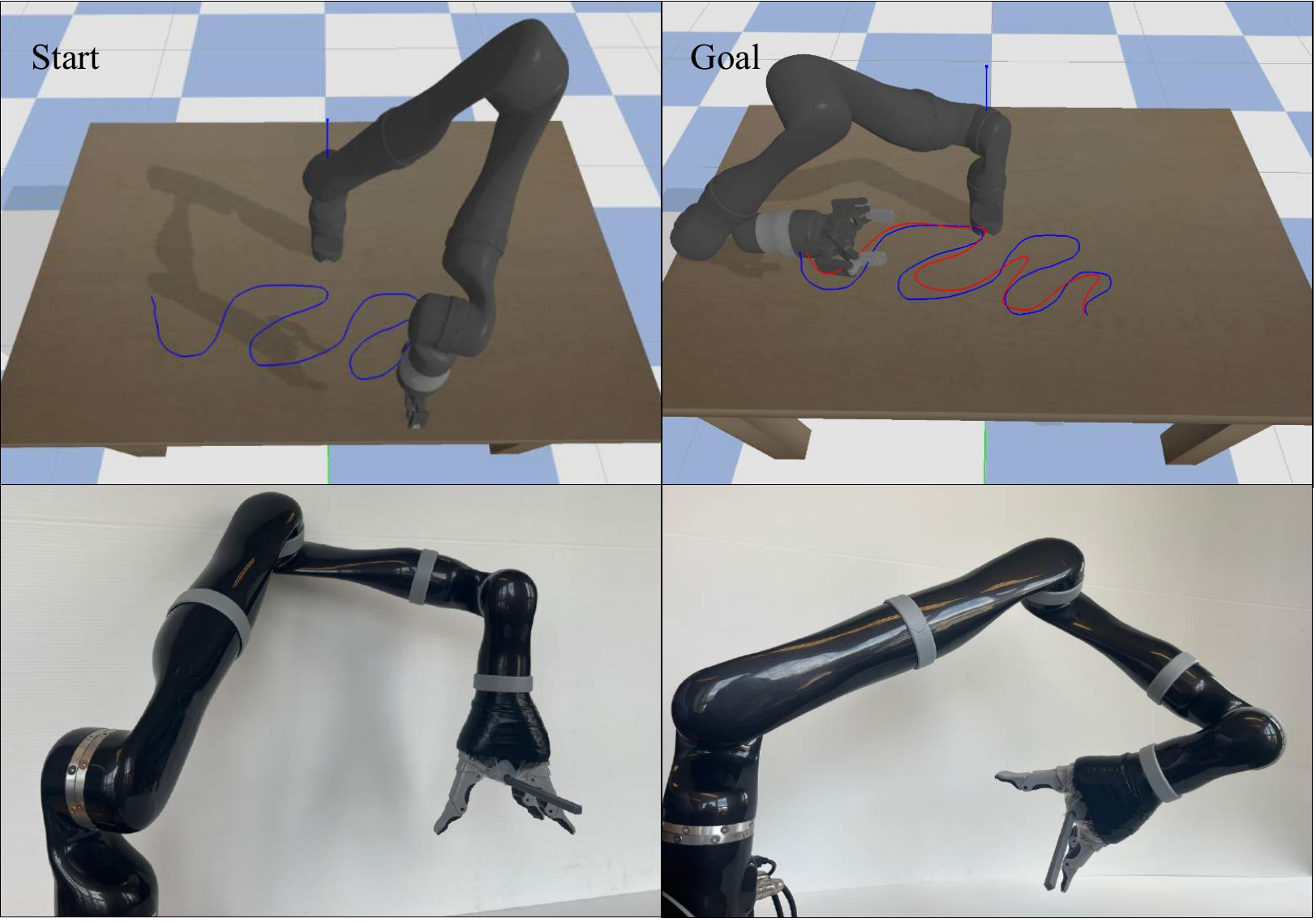}
    \caption{Policy deployment in simulation (top) and real-world Kinova Jaco2 arm (bottom) scenarios. The policy is trained on a single Messy Snake demonstration. The simulation rollouts (red) are close to the expert's demonstration (blue), and are similar to the rollouts in \Cref{fig:rollout_messy_snake}.}
    \label{fig:pybullet_realworld_exps}
\end{figure}

In \Cref{fig:rollout_messy_snake}, we compare the performance of our method against the baselines trained with the Messy Snake data in SE(3). We utilize the DTW metric to evaluate the policy rollouts against the original trajectories. Next, we deploy the policy trained with \ours{} on the simulated arm in PyBullet. \Cref{fig:pybullet_realworld_exps} offers an overview of policy deployment in simulation trained on the Messy Snake data.

Finally, we deploy the learned policy directly on a similar real-world manipulator platform.
Recordings of experiments conducted with the Kinova Jaco2 arm are part of the supplementary video. 
Upon viewing the footage, it is evident that the robot reproduces the same motion as the simulation. 
The smooth sim-to-real transfer is made possible by the presence of Lyapunov stability conditions.
The learned policy is robust to model error, since the Lyapunov condition ensures that the robot consistently progresses toward the target.

\section{Discussion}
\noindent The experiments demonstrate that employing \ours{} to model a policy results in improved predictability and safety when recovering from states far beyond those covered in the original demonstrations. \ours{} proves to be more effective by utilizing expressive and scalable neural architecture in comparison to the state of the art. Consequently, \ours{} can acquire highly nonlinear stable policies in larger state spaces with reduced computational overhead. Our approach also reduces the restrictive assumptions regarding the class of Lyapunov functions to convexity, while the adopted projection technique ensures a smooth training process on a variety of expert data, as evident by the experiments.

While SNDS offers global convergence and predictability, there remains a concern regarding physically impossible trajectories for robots to replicate, particularly when considering manipulators and their joint or torque limits. Future enhancements could address this by incorporating safe regions using control barrier functions. Moreover, it is worth noting that modeling the Lyapunov candidate as a strictly convex function is not an essential requirement. Future research could explore invertible transformations to ease this restriction. Additionally, integrating obstacle avoidance and joint constraints into our formulation are promising paths to explore beyond the current scope. More ambitious extensions might delve into the implications of utilizing SNDS policies in reinforcement learning or training and deploying stable policies on legged robots, especially for gait control.

\section{Conclusion}
\label{sec:conclusion}

\noindent We outlined a training process to learn expressive neural policies through imitating expert demonstrations. We effectively enforce global stability throughout the training process by adhering to the conditions of Lyapunov stability theory. This ensures that the resulting policy reliably converges to a predetermined target, regardless of initial conditions, velocity and time variations, or unexpected perturbations encountered during deployment. Our theoretical findings are accompanied by simulation and real-world benchmarking against best-performing methods in the field.

\section{Reproducability}
\label{sec:appendix}

\noindent \textbf{Network architecture and hyperparameters.} Our approach employs a 4-hidden layer feed-forward neural network for policy, with Leaky ReLU activations and layer sizes of 2-256-256-128-128-2. For the Lyapunov function, we utilize an ICNN architecture~\cite{icnn_paper} with 3 hidden layers of 2-128-128-128-1 nodes along with Leaky ReLU and Soft plus activations as required by the original design. We set $\gamma_i = \frac{1}{2^i}$ and $N_w = 2$ to limit the cost function's horizon. Other parameters, e.g., learning rate and batch size, are selected and explained in our code repository. We use grid-search to pick the hyperparameters if computationally feasible. 
\vspace{0.1cm}

\noindent \textbf{Datasets specifications.} The Handwriting Dataset~\cite{handwriting_lasa} contains 30 sets of 2D handwriting motions recorded from a Tablet-PC via user input. There are seven demonstrations per motion, starting from slightly different initial positions but ending at (0, 0). Each demonstration encompasses position (2 × 1000) as learning features and velocity (2 × 1000) as labels. We pick the following representative set of motion for our experiments: \textit{Sine}, \textit{P}, \textit{Worm}, \textit{G}, \textit{N}, \textit{DBLine}, \textit{C}, \textit{Angle}. Note that the \textit{Messy Snake} motion is not a part of this dataset. The Messy Snake data, gathered in \cite{figueroa18a}, comprises three demonstrations with various sample sizes per demonstration, totaling to (2 × 1592), and we sample a single demonstration for policy learning and deployment.
\vspace{0.1cm}

\noindent \textbf{Computational resources.} We utilize a single machine with {NVIDIA GeForce RTX 4050} GPU, {Intel Core i7-13620H} CPU, and {32 GB DDR4} RAM.  
\vspace{0.1cm}

\noindent \textbf{Codebase and reproduction.} Consult \texttt{README.md} in our GitHub repository \href{https://www.github.com/aminabyaneh/stable-imitation-policy}{github.com/aminabyaneh/stable-imitation-policy} to access \ours{} experiments.

\section*{Acknowledgment}
\noindent
We express our heartfelt gratitude to Anya Forestell for her invaluable hardware support during our experiments.

\clearpage
\addcontentsline{toc}{section}{References}
\bibliographystyle{IEEEtran}
\bibliography{references}

\begin{thebibliography}{10}
\providecommand{\url}[1]{#1}
\csname url@rmstyle\endcsname
\providecommand{\newblock}{\relax}
\providecommand{\bibinfo}[2]{#2}
\providecommand\BIBentrySTDinterwordspacing{\spaceskip=0pt\relax}
\providecommand\BIBentryALTinterwordstretchfactor{4}
\providecommand\BIBentryALTinterwordspacing{\spaceskip=\fontdimen2\font plus
\BIBentryALTinterwordstretchfactor\fontdimen3\font minus
  \fontdimen4\font\relax}
\providecommand\BIBforeignlanguage[2]{{%
\expandafter\ifx\csname l@#1\endcsname\relax
\typeout{** WARNING: IEEEtran.bst: No hyphenation pattern has been}%
\typeout{** loaded for the language `#1'. Using the pattern for}%
\typeout{** the default language instead.}%
\else
\language=\csname l@#1\endcsname
\fi
#2}}

\bibitem{schaal1999imitation}
S.~Schaal, ``Is imitation learning the route to humanoid robots?'' \emph{Trends
  in cognitive sciences}, vol.~3, no.~6, pp. 233--242, 1999.

\bibitem{hussein2017imitation}
A.~Hussein, M.~M. Gaber, E.~Elyan, and C.~Jayne, ``Imitation learning: A survey
  of learning methods,'' \emph{ACM Computing Surveys (CSUR)}, vol.~50, no.~2,
  pp. 1--35, 2017.

\bibitem{figueroa2022locally}
N.~Figueroa and A.~Billard, ``Locally active globally stable dynamical systems:
  Theory, learning, and experiments,'' \emph{The International Journal of
  Robotics Research}, vol.~41, no.~3, pp. 312--347, 2022.

\bibitem{ds_billard}
M.~Hersch, F.~Guenter, S.~Calinon, and A.~Billard, ``Dynamical system
  modulation for robot learning via kinesthetic demonstrations,'' \emph{IEEE
  Transactions on Robotics}, vol.~24, no.~6, pp. 1463--1467, 2008.

\bibitem{khansari2011learning}
S.~M. Khansari-Zadeh and A.~Billard, ``Learning stable nonlinear dynamical
  systems with gaussian mixture models,'' \emph{IEEE Transactions on Robotics},
  vol.~27, no.~5, pp. 943--957, 2011.

\bibitem{khansari2014learning}
S.~M. Khansari{-}Zadeh and A.~Billard, ``Learning control {L}yapunov function
  to ensure stability of dynamical system-based robot reaching motions,''
  \emph{Robotics and Autonomous Systems}, vol.~62, no.~6, pp. 752--765, 2014.

\bibitem{figueroa18a}
N.~Figueroa and A.~Billard, ``A physically-consistent bayesian non-parametric
  mixture model for dynamical system learning,'' in \emph{2nd Annual Conference
  on Robot Learning}, 2018, pp. 927--946.

\bibitem{neumann2015learning}
K.~Neumann and J.~J. Steil, ``Learning robot motions with stable dynamical
  systems under diffeomorphic transformations,'' \emph{Robotics and Autonomous
  Systems}, vol.~70, pp. 1--15, 2015.

\bibitem{ravichandar2017learning}
H.~Ravichandar, I.~Salehi, and A.~Dani, ``Learning partially contracting
  dynamical systems from demonstrations,'' in \emph{Conference on Robot
  Learning}.\hskip 1em plus 0.5em minus 0.4em\relax PMLR, 2017, pp. 369--378.

\bibitem{khadivar2021learning}
F.~Khadivar, I.~Lauzana, and A.~Billard, ``Learning dynamical systems with
  bifurcations,'' \emph{Robotics and Autonomous Systems}, vol. 136, p. 103700,
  2021.

\bibitem{abyaneh2023plyds}
A.~Abyaneh and H.-C. Lin, ``Learning {L}yapunov-stable polynomial dynamical
  systems through imitation,'' in \emph{7th Annual Conference on Robot
  Learning}, 2023.

\bibitem{chang2019neurallyapunov}
Y.-C. Chang, N.~Roohi, and S.~Gao, ``Neural {L}yapunov control,''
  \emph{Advances in neural information processing systems}, vol.~32, 2019.

\bibitem{rana2020euclideanizing}
M.~A. Rana, A.~Li, D.~Fox, B.~Boots, F.~Ramos, and N.~Ratliff, ``Euclideanizing
  flows: Diffeomorphic reduction for learning stable dynamical systems,'' in
  \emph{Learning for Dynamics and Control}.\hskip 1em plus 0.5em minus
  0.4em\relax PMLR, 2020, pp. 630--639.

\bibitem{zhang2022riemaniandiffeomorphism}
J.~Zhang, H.~B. Mohammadi, and L.~Rozo, ``Learning riemannian stable dynamical
  systems via diffeomorphisms,'' in \emph{6th Annual Conference on Robot
  Learning}, 2022.

\bibitem{ho2016generative}
J.~Ho and S.~Ermon, ``Generative adversarial imitation learning,''
  \emph{Advances in neural information processing systems}, vol.~29, p.
  4572–4580, 2016.

\bibitem{airl_imitation}
J.~Fu, K.~Luo, and S.~Levine, ``Learning robust rewards with adversarial
  inverse reinforcement learning,'' in \emph{International Conference on
  Learning Representations}, 2018.

\bibitem{dai2021lyapunov}
H.~Dai, B.~Landry, L.~Yang, M.~Pavone, and R.~Tedrake, ``Lyapunov-stable
  neural-network control,'' in \emph{Robotics: Science and Systems}, 2021.

\bibitem{coulombe2022}
A.~Coulombe and H.-C. Lin, ``Generating stable and collision-free policies
  through {L}yapunov function learning,'' in \emph{International Conference on
  Robotics and Automation}, 2023, pp. 3037--3043.

\bibitem{abbeel2004apprenticeship}
P.~Abbeel and A.~Y. Ng, ``Apprenticeship learning via inverse reinforcement
  learning,'' in \emph{International conference on Machine learning}, 2004,
  p.~1.

\bibitem{ziebart2008maximumentropy}
B.~D. Ziebart, A.~L. Maas, J.~A. Bagnell, A.~K. Dey, \emph{et~al.}, ``Maximum
  entropy inverse reinforcement learning.'' in \emph{Association for the
  Advancement of Artificial Intelligence}, vol.~8, 2008, pp. 1433--1438.

\bibitem{devaney2021introduction}
R.~L. Devaney, \emph{An introduction to chaotic dynamical systems}.\hskip 1em
  plus 0.5em minus 0.4em\relax CRC press, 2021.

\bibitem{icnn_paper}
B.~Amos, L.~Xu, and J.~Z. Kolter, ``Input convex neural networks,'' in
  \emph{International Conference on Machine Learning}.\hskip 1em plus 0.5em
  minus 0.4em\relax PMLR, 2017, pp. 146--155.

\bibitem{manek2019learning}
J.~Z. Kolter and G.~Manek, ``Learning stable deep dynamics models,''
  \emph{Advances in neural information processing systems}, vol.~32, 2019.

\bibitem{handwriting_lasa}
S.~M. Khansari-Zadeh and A.~Billard, ``L{A}{S}{A} {H}andwriting {D}ataset,''
  \url{https://cs.stanford.edu/people/khansari/download.html\#SEDS\_reference},
  2011.

\bibitem{bruveris2016srvf}
M.~Bruveris, ``Optimal reparametrizations in the square root velocity
  framework,'' \emph{SIAM Journal on Mathematical Analysis}, vol.~48, no.~6,
  pp. 4335--4354, 2016.

\bibitem{adam_optimizer}
D.~P. Kingma and J.~Ba, ``Adam: A method for stochastic optimization,'' in
  \emph{International Conference on Learning Representations}, 2015.

\bibitem{pomerleau1988behavioralcloning}
D.~A. Pomerleau, ``Alvinn: An autonomous land vehicle in a neural network,'' in
  \emph{Advances in neural information processing systems}, vol.~1, 1988, pp.
  305--313.

\bibitem{torabi2018behavioral}
F.~Torabi, G.~Warnell, and P.~Stone, ``Behavioral cloning from observation,''
  in \emph{Proceedings of the 27th International Joint Conference on Artificial
  Intelligence}, 2018, pp. 4950--4957.

\bibitem{keogh2005dtw}
E.~Keogh and C.~A. Ratanamahatana, ``Exact indexing of dynamic time warping,''
  \emph{Knowledge and information systems}, vol.~7, pp. 358--386, 2005.

\end{thebibliography}

\clearpage

\onecolumn
\section*{\textbf{Appendix}}

\subsection{Efficient trajectory generation}
\label{sec:efficient_traj_gen}
\noindent \Cref{sec:srvf_training_loss} introduces a novel loss that penalizes discrepancies not only in the action (velocity) space but also in the state space. Penalizing the policy in the state space is intuitive: by using the forward Euler method, we can efficiently estimate the loss directly in the state space. Consider a scenario where all but one of the velocities are perfectly imitated. Even a single discrepancy in velocity can lead to a significant deviation in the state space, resulting in a substantial loss, while the corresponding loss in the action space remains minimal. 
\vspace{5pt}

To address this issue, the loss formulation relies on short-horizon, efficient forward simulation of the policy to obtain trajectories in the state space. In this section, we briefly discuss two methods for generating such trajectories.
\vspace{5pt}

\noindent \textbf{Direct solution. } Remember the following equation in \Cref{sec:srvf_training_loss}: 
$$\sum_{i \in [N_w]} \gamma_i \mathop{\mathbb{E}} \Big[\big(\state^{\demonstration}[\sample + i - 1] +  \dsStableSym (\state^{\demonstration}[\sample + i - 1]) \Delta t - \state^\demonstration[\sample + i]\big)^2\Big],$$ 

\noindent where we can simply calculate a limited number of Euler steps to optimize the policy in the state space. This can be readily implemented by solving the DS policy for a given initial state using the Euler method. By selecting a sufficiently small step size, $\Delta t$, and an appropriate solution horizon, $N_w$, it is possible to train accurate policies while maintaining acceptable computation speed. Importantly, when the policy is modeled with automatic differentiation tools, the entire solution remains differentiable. Additionally, with a small horizon, the memory consumption and computational complexity are manageable. This is \textbf{our chosen approach} due to its efficiency and ease of implementation.

\vspace{5pt}

\noindent \textbf{Fixed-point differentiation. } Another approach involves solving the DS policy for a fixed horizon and given initial condition using the implicit function theorem. This method works by treating the entire trajectory as the solution of an implicit equation governing the system's dynamics—in this case, our DS policy, $\dsStableSym$. Unlike traditional numerical methods (like Euler's method), which explicitly compute each step of the trajectory, the implicit function theorem enables backpropagation through the entire trajectory by implicitly differentiating the final state with respect to the initial conditions and model parameters. This allows for efficient gradient computation in a memory-friendly way, as only the final state and system dynamics need to be maintained, avoiding the storage of all intermediate states. Solving the DS policy through the implicit function theorem offers a powerful alternative to explicit numerical methods, particularly when managing memory and computational resources. However, we observed that for small horizons (a few forward Euler steps), the numerical method is computationally faster, and the additional memory consumption is negligible. Therefore, while the implicit function approach is valuable for more complex or extended horizons, the simplicity and efficiency of the numerical method make it preferable for short horizons in our use case.

\subsection{Smoothness and numerical stability}
\label{sec:smooth_reg}
\noindent Non-smooth behavior can be observed in the policy, and is attributed to the use of a non-differentiable activation function with the projection operation, which introduces discontinuities in the gradient at zero. Even if the activation function is smooth, the denominator in \Cref{eq:projection_formula} can become infinitesimal, essentially breaking the training process. To address this issue, we propose substituting ReLU with a smooth approximation, such as Softplus, or a customized activation suggested in \cite{manek2019learning}. Once the activation function is smooth and differentiable, incorporating a small regularization term in the denominator of the projection operation in \Cref{eq:projection_formula} helps to prevent abrupt changes in the output by stabilizing the computation when the denominator approaches zero. 
\vspace{5pt}

The smoothness of all trajectories is of utmost importance, especially when imitating expert behavior in the real-world, where the robot is expected to follow smooth and reliable trajectories in the entire state space. Further refinements can be achieved by applying gradient clipping, which controls extreme gradient values that could otherwise contribute to non-smoothness. By limiting the gradient magnitude, we ensure that the function behaves more consistently across different input values. One last precaution is to ensure that the underlying Lyapunov candidate is approximated smoothly, possibly by adjusting its architecture to a Lipschitz bounded network. 

\vspace{5pt}

\subsection{Additional experiments}
\label{sec:more_exps}
\begin{figure}
    \centering
    \includegraphics[width=1\linewidth]{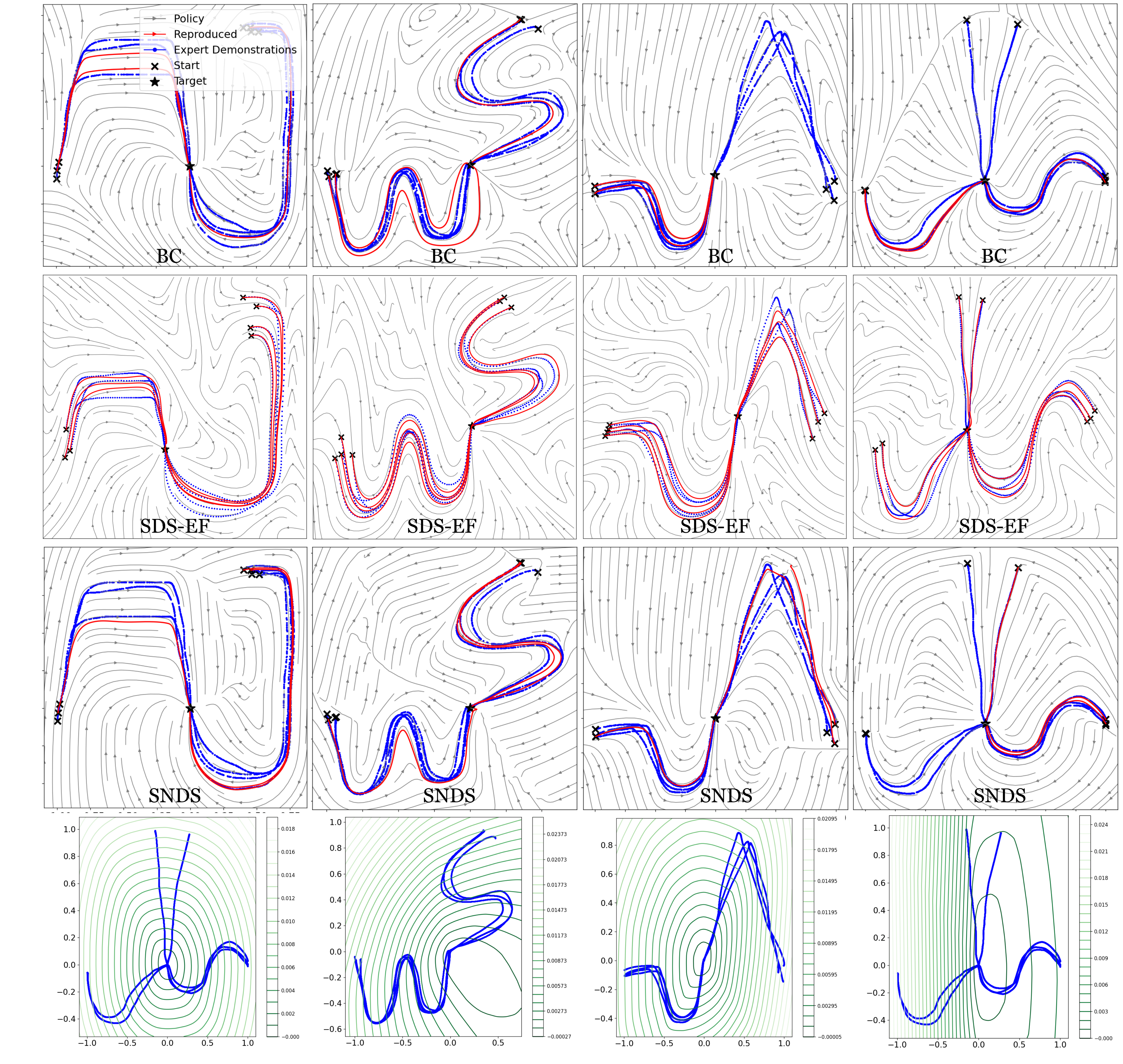}
    \caption{\ours{} is compared against with SDS-EF (stable) and BC (unstable) learning methods. This experiment uses the multimodel motions in the handwriting motions. The trained Lyapunov function is displayed in the last row. }
    \label{fig:baseline_multimodels}
\end{figure}

\noindent For additional experiments, we pick the multimodel motions in the handwriting dataset. These motions consist of multiple motions converging to a global equilibrium from different directions, and with initial states that are far apart. \Cref{fig:baseline_multimodels} presents a thorough collection of empirical results, as \ours{} is compared against two other baselines: SDS-EF and BC. 

\end{document}